\documentclass[10pt,twocolumn,letterpaper]{article}

\usepackage[pagenumbers]{cvpr} 

\usepackage[pagebackref,breaklinks,colorlinks]{hyperref}
\usepackage{float}
\usepackage{afterpage}

\def\paperID{12422} 
\def\confName{CVPR}
\def\confYear{2026}

\title{ SAM Guided Semantic and Motion Changed Region Mining for Remote Sensing Change Captioning }

\author{Futian Wang$^{1}$, Mengqi Wang$^{1}$, Xiao Wang$^{1}$\thanks{Corresponding Author: Xiao Wang, Haowen Wang}, Haowen Wang$^{1}$, Jin Tang$^{1}$ \\ 
${^1}${School of Computer Science and Technology, Anhui University, Hefei 230601, China} \\ 
\textit{\{wft, xiaowang, wanghaowen, tangjin\}@ahu.edu.cn}, e24301148@stu.ahu.edu.cn 
}

\begin{document}
\maketitle

\begin{abstract}
Remote sensing change captioning is an emerging and popular research task that aims to describe, in natural language, the content of interest that has changed between two remote sensing images captured at different times. 
Existing methods typically employ CNNs/Transformers to extract visual representations from the given images or incorporate auxiliary tasks to enhance the final results, with weak region awareness and limited temporal alignment. 
To address these issues, this paper explores the use of the SAM (Segment Anything Model) foundation model to extract region-level representations and inject region-of-interest knowledge into the captioning framework. Specifically, we employ a CNN/Transformer model to extract global-level vision features, leverage the SAM foundation model to delineate semantic- and motion-level change regions, and utilize a specially constructed knowledge graph to provide information about objects of interest. These heterogeneous sources of information are then fused via cross-attention, and a Transformer decoder is used to generate the final natural language description of the observed changes. Extensive experimental results demonstrate that our method achieves state-of-the-art performance across multiple widely used benchmark datasets. The source code of this paper will be released on \url{https://github.com/Event-AHU/SAM_ChangeCaptioning} 
\end{abstract}

\section{Introduction}

The goal of the remote sensing change captioning task~\cite{liu2022remote} is to express, in natural language, the changes in objects of interest between two given remote sensing images captured at different times. This task can be widely applied in urban planning and land-use monitoring, disaster emergency response, environmental and ecological conservation, as well as military and security surveillance. Although some progress has been made in recent years, significant challenges remain, including difficulties in multi-temporal image alignment and registration, the complexity and diversity of change semantics, insufficient fine-grained semantic understanding, and misalignment between language generation and visual changes. Thus, high-quality remote sensing image change captioning remains an unsolved problem.

\begin{figure}
\centering
\includegraphics[width=0.85\linewidth]{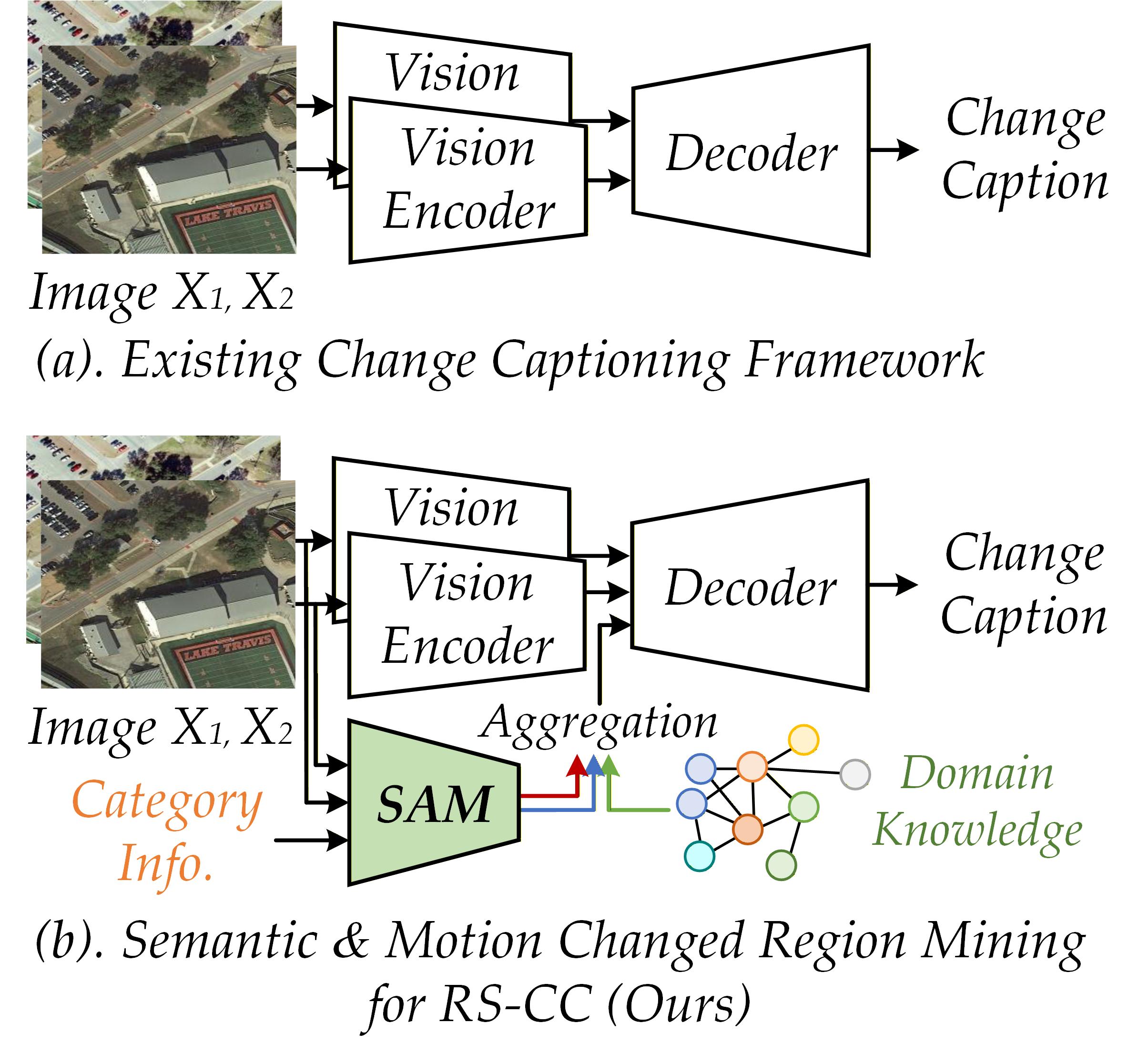}
\caption{Comparison between (a) existing change captioning framework and (b) our newly proposed SAM guided semantic and motion changed region mining for RS-CC.}
\label{fig:firstIMG}
\end{figure}

Current work is based on CNN~\cite{kim2014convolutional}, LSTM~\cite{graves2012long}, and Transformer~\cite{vaswani2017attention} networks, which have significantly advanced the research, as shown in Fig.~\ref{fig:firstIMG}. Specifically, Daudt et al.~\cite{daudt2018fully} proposed a fully convolutional network (FCN) architecture extended into a Siamese branch structure to learn pixel-level change maps from bi-temporal images. Lv et al.~\cite{lv2023multiscale} proposed a CNN model based on a UNet backbone, integrating multi-scale attention and change-gradient modules to enhance change detection accuracy. Papadomanolaki et al.~\cite{papadomanolaki2019detecting} proposed a hybrid model combining FCN for spatial feature extraction and LSTM for modeling temporal dependencies, enhancing urban change detection from multitemporal Sentinel-2 data. Chen et al.~\cite{chen2021remote} proposed a Transformer-based framework for bi-temporal images, using semantic tokens to model spatio-temporal context for improved change detection. Bandara and Patel~\cite{bandara2022transformer} designed a Transformer-based Siamese architecture, with two branches processing bi-temporal inputs, leveraging multi-scale long-range attention to enhance detail perception in change detection.

Although these works have greatly advanced the field, we believe they are still limited in the following aspects: 
1). Existing models primarily focus on employing general or hybrid network architectures, such as CNNs~\cite{kim2014convolutional} or Transformers~\cite{vaswani2017attention}, to extract visual representations from given remote sensing images. However, few studies have considered incorporating semantic and temporal motion-related changes to provide richer contextual details for enhancing the final textual descriptions. 
2). Common object categories in the scene and their relationships play a crucial role in remote sensing change description, for example, objects of interest may include \textit{buildings, roads, vegetation, bridges}, etc. However, mainstream models fail to explicitly exploit and incorporate this information, often causing the models to focus on insignificant, fine-grained changes and significantly degrading the semantic accuracy of the final results. 
Thus, it is natural to raise the following question: ``\textit{How can we effectively mine fine-grained change regions and distinguish the truly relevant changes from them?}"

In this paper, we propose a novel remote sensing change captioning framework that synergistically combines pre-trained foundation models, multi-level visual representation, and structured domain knowledge. Our key innovation lies in the integration of the Segment Anything Model (SAM)~\cite{kirillov2023segment}, a powerful vision foundation model, as a region-aware change analyzer. Unlike conventional methods that treat images holistically, we leverage SAM to explicitly delineate regions exhibiting semantic-level and motion-level changes between image pairs. This enables our model to answer not only “\textit{where did the change occur?}” but also “\textit{what changed?}” and crucially, “\textit{is this change relevant or of interest?}”, a capability essential for generating accurate and meaningful descriptions. In addition, our framework goes further by incorporating a specially constructed knowledge graph that encodes prior information about objects commonly involved in meaningful scene changes. This knowledge acts as a semantic prior, guiding the captioning process with contextual cues that pure visual analysis might miss. The heterogeneous signals are then effectively fused through a cross-attention mechanism, allowing the model to dynamically weigh visual evidence against semantic expectations. Finally, a Transformer-based decoder is adopted to generate fluent and precise natural language captions. Our work bridges the gap between low-level change detection and high-level semantic interpretation, offering a more robust and interpretable solution to this complex multi-modal task. An overview of our proposed framework can be found in Fig.~\ref{fig:framework}.

To sum up, the main contributions of this paper can be summarized as follows: 

1). We propose a novel remote sensing change captioning framework that leverages the Segment Anything Model to explicitly identify semantic- and motion-level change regions, enabling accurate localization of what changed and whether it matters.

2). We construct a semantic knowledge graph using large language models and integrate it into the captioning pipeline as a contextual prior, enhancing semantic coherence and relevance of the generated descriptions.

3). Extensive experiments on three widely used benchmark datasets fully demonstrate state-of-the-art performance, validating the effectiveness of our approach.

\section{Related Works} 

\subsection{Remote Sensing Change Captioning} 
Remote Sensing Change Captioning (RS-CC) is one of the core tasks in the field of vision-language understanding for remote sensing time-series images. 
Jhamtani et al.~\cite{jhamtani2018learning} propose the change captioning (CC) task and introduce a new dataset along with a Siamese CNN-RNN architecture to capture significant differences between similar images. Subsequently, two small-scale RSICC datasets, namely LEVIR CCD and Dubai CCD, are released, focusing on remote sensing change captioning. To further advance the research, Liu et al.~\cite{liu2022remote} provide a new large-scale RSICC dataset, LEVIR-CC, and build an RSICC network based on the Transformer architecture, effectively demonstrating the adaptability of Transformer structures in RSICC tasks. Chang and Ghamisi~\cite{chang2023changes} integrate a Siamese CNN with an attention-based Transformer encoder-decoder to dynamically locate change regions. Cai et al.~\cite{cai2023interactive} introduce ICT-Net, employing cross-gated attention and adaptive fusion to enhance change representation. Sun et al.~\cite{sun2024lightweight} propose a Sparse-Focused Transformer (SFT) that reduces computational cost while maintaining competitive performance. Zhou et al.~\cite{zhou2024single} develop a Single-stream Feature Extraction Network (SEN) with contrastive pretraining and cross-attention modules to improve efficiency. Zhu et al.~\cite{zhu2025change3d} introduce Change3D, which treats bi-temporal images as video frames to achieve unified modeling of change detection and captioning. Sun et al.~\cite{sun2025mask} introduce a diffusion-based data distribution learning framework with frequency-domain noise filtering to overcome overfitting and improve model generalization. Different from these works, we propose a method that combines the SAM and a knowledge graph to identify change regions and fuse visual and semantic information, enhancing the accuracy and consistency of change descriptions.

\subsection{Segment Anything Model}   
The Segment Anything Model (SAM)~\cite{kirillov2023segment} is a large-scale pre-trained model with strong zero-shot generalization. Inspired by the prompt strategy in NLP, SAM segments diverse targets based on various prompt types, including points, bounding boxes, masks, and text. Following SAM’s release, numerous studies propose improvements and domain-specific adaptations to enhance its performance and generalization~\cite{yue2024surgicalsam, liu2023explicit, liu2023matcher, xue2024adapting, zhong2024convolution}. 

Cheng et al.~\cite{cheng2023sammed2d} propose SAM-Med2D, which adapts SAM for medical image segmentation through large-scale fine-tuning and diverse prompting strategies. Chen et al.~\cite{chen2024rsprompter} propose RSPrompter, combining SAM with semantic category information to automatically generate prompts for semantic instance segmentation in remote sensing images. Ravi et al.~\cite{ravi2024sam} introduce SAM2, which employs a data engine and streaming memory transformer for real-time image and video segmentation, improving both accuracy and efficiency. Grounded SAM~\cite{ren2024grounded} integrates Grounding DINO~\cite{liu2024grounding} with SAM to improve open-vocabulary image segmentation and extend its applicability to more controllable tasks. Inspired by these works, in this paper, we adopt SAM to analyze the given remote sensing images to better locate meaningful changed regions.

\subsection{Knowledge Graph} 

Knowledge Graphs (KGs), first introduced by Google in 2012, represent entities and relations through graph structures for knowledge organization and reasoning. With the rise of deep learning and NLP, neural models such as BiLSTM~\cite{zhang2015bidirectional}, CNN~\cite{kim2014convolutional}, and Transformer~\cite{vaswani2017attention} become dominant in entity and relation extraction. Recently, large language models (LLMs) open new directions for KG construction and application by efficiently extracting, completing, and reasoning over knowledge. The integration of LLMs with KGs, either by using KGs as external memory or by leveraging LLMs to expand and refine them, becomes a major research trend. Several frameworks explore different forms of KG–LLM integration. Kim et al.~\cite{kim2023kg} propose KG-GPT, combining sentence segmentation, graph retrieval, and inference for fact verification and question answering. Shu et al.~\cite{shu2024knowledge} introduce KG-LLM, converting KG structures into natural language prompts for multi-hop link prediction. Mo et al.~\cite{mo2025kggen} present KGGen, which improves extracted graphs through iterative clustering and relation grouping.

Graph Neural Networks (GNNs), such as R-GCN~\cite{schlichtkrull2018modeling} and CompGCN~\cite{vashishth2019composition}, effectively combine structural and semantic information for KG representation learning and reasoning. Recent research integrates KGs with Natural Language Generation (NLG) to improve factuality and interpretability in descriptive and report generation. Soman et al.~\cite{soman2024biomedical} propose KG-RAG for biomedical text generation; Wu et al.~\cite{wu2024text} introduce a multimodal KG-driven diagnostic text generator; and Wang et al.~\cite{wang2025r2genkg} propose R2GenKG, which builds a large-scale multimodal medical KG (M3KG) using GPT-4o and R-GCN to generate high-quality X-ray reports. Different from these works, we use a relation-aware knowledge graph to bridge visual and semantic differences in dual-time remote sensing images, enhancing temporal reasoning and caption generation.

\section{Methodology} 

\begin{figure*}
\centering
\includegraphics[width=1\textwidth]{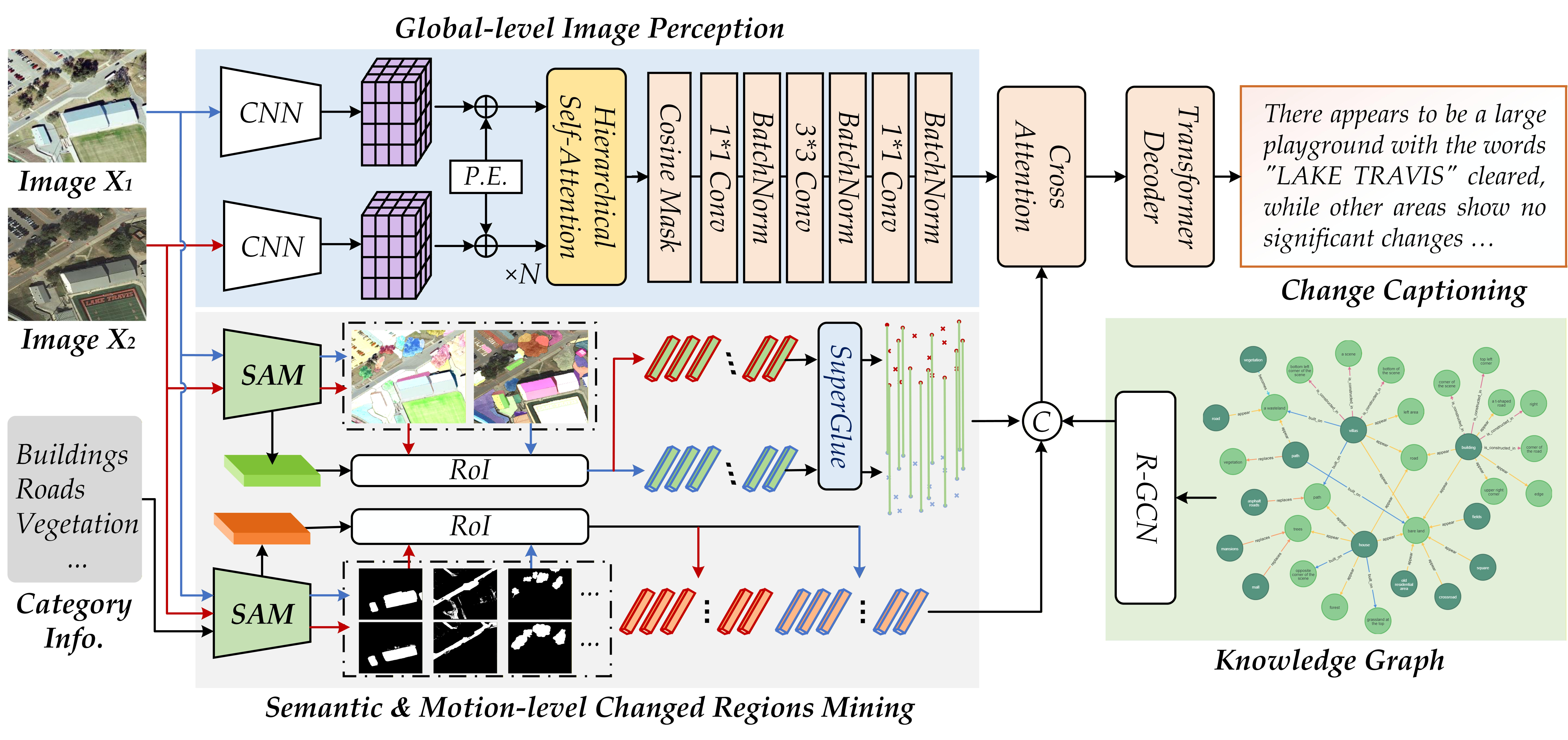}
\caption{An illustration of our proposed remote sensing change captioning by SAM guided semantic, motion changed region mining and knowledge graph.} 
\label{fig:framework} 
\end{figure*}

\subsection{Overview} 

As illustrated in Fig.~\ref{fig:framework}, we propose Segment-Assisted Graph-Enhanced Change Captioner~(SAGE-CC), a framework for remote sensing change captioning from bi-temporal image pairs acquired at two epochs.
Specifically, 
the Bi-Temporal Scene Consistency Encoder~(Sec.~\ref{sec:btsce}) extracts shared multi-scale features and distills explicit cross-time consistency priors, offering robust global context to support subsequent modules. 
The SAM-Guided Change Region Mining~(Sec.~\ref{sec:scrl}) leverages SAM to jointly perform motion-level change localization of pixel-accurate changed regions and prompt-guided aggregation of semantics into category-aware region representations. 
Moreover, the Remote-Sensing Change Graph Reasoner~(Sec.~\ref{sec:rscgr}) constructs a compact knowledge graph from training captions and employs relation-aware message passing to generate domain-specific priors on key entities and typical change patterns. 
Finally, the Change-Aware Language Generator~(Sec.~\ref{sec:calg}) integrates global scene priors, localized region representations, and knowledge graph priors, producing concise and faithful change descriptions that clearly articulate \emph{where}, \emph{what}, and \emph{how} the scene changes. 
Training objectives and optimization details are provided in Sec.~\ref{sec:calg}.

\subsection{Bi-Temporal Scene Consistency Encoder}
\label{sec:btsce}

This module generates a global scene embedding and explicit pixel-wise consistency and change priors from a bi-temporal image pair.

\noindent\textbf{Spatial Feature Backbone.}
Given a bi-temporal image pair $X=(X_1,X_2)$, a Siamese ResNet-101~\cite{2016Deep} with shared parameters extracts feature maps $F_1,F_2\in\mathbb{R}^{h\times w\times C}$.
To preserve spatial structure, a 2D positional embedding $P\in\mathbb{R}^{h\times w\times C}$ is added to each stream:
\begin{equation}
F_i^{(0)} = F_i + P,\qquad i\in\{1,2\},
\end{equation}
producing coordinate-aware features in a unified representation space.

\noindent\textbf{Cross-Time Consistency Modeling.}
To enhance temporal reasoning, we employ $N$ lightweight attention blocks that jointly refine intra-temporal semantics and highlight inter-temporal discrepancies. 
In each block $l$, every stream first applies multi-head self-attention (SA) and then cross-attends (CA) to the concatenated bi-temporal context:
\begin{align}
H_i^{(l)} &= \mathrm{SA}\!\big(F_i^{(l-1)}\big) + F_i^{(l-1)},\\
F_i^{(l)} &= \mathrm{CA}\!\big(H_i^{(l)},[H_1^{(l)},H_2^{(l)}]\big) + H_i^{(l)},\quad i\in\{1,2\}.
\end{align}
After $N$ blocks we obtain temporally aligned features $F_1^{(N)},F_2^{(N)}\in\mathbb{R}^{h\times w\times C}$ and compute a pixel-wise cosine similarity:
\begin{equation}
S(u,v)=\frac{\langle F_1^{(N)}(u,v,:),\,F_2^{(N)}(u,v,:)\rangle}
{\|F_1^{(N)}(u,v,:)\|_2\,\|F_2^{(N)}(u,v,:)\|_2}.
\end{equation}
The similarity map $S$ is normalized to obtain a \emph{consistency prior} $C_{\mathrm{map}}=(S+1)/2\in[0,1]^{h\times w}$, and its complement is defined as the \emph{change prior} $M_{\mathrm{map}}=1-C_{\mathrm{map}}$. 
We then fuse features under the guidance of these priors:
\begin{equation}
\label{eq:prior-fusion}
\begin{aligned}
G
= \mathrm{concat}\Big(
&F_1^{(N)}\odot C_{\mathrm{map}},\;
F_2^{(N)}\odot C_{\mathrm{map}},\\
&|F_1^{(N)}-F_2^{(N)}|\odot M_{\mathrm{map}},\;
M_{\mathrm{map}}\Big),
\end{aligned}
\end{equation}
which is processed by a shallow Conv–BN–ReLU head followed by a $1{\times}1$ projection to yield
\begin{equation}
E_{\mathrm{img}}=\phi_{\theta}(G),\qquad E_{\mathrm{img}}\in\mathbb{R}^{h\times w\times C_e}.
\end{equation}
The resulting global embedding $E_{\mathrm{img}}$ and priors $\{C_{\mathrm{map}},M_{\mathrm{map}}\}$ are forwarded to subsequent modules for downstream reasoning.

\subsection{SAM-Guided Change Region Mining}
\label{sec:scrl}

SAM-Guided Change Region Mining leverages SAM~\cite{kirillov2023segment} to localize pixel-accurate, motion-level change regions and to aggregate semantic-level change features into category-aware region representations from the bi-temporal pair.

\subsubsection{Motion-Level Change Localization}
\label{sec:scrl-disc}

Given each image $X_i\,(i\!\in\!\{1,2\})$, SAM first generates a set of instance masks $\mathcal{M}_i=\{m_{i,n}\}_{n=1}^{N_i}$ and exposes an intermediate dense feature map $S_i\in\mathbb{R}^{h\times w\times C_s}$ from its frozen image encoder.
To select reliable and complete region proposals, we filter candidate masks based on confidence scores, region area, non-maximum suppression, and a minimum-area threshold.


Each mask is converted into a bounding box $r_{i,n}$, and corresponding region features are extracted using RoIAlign:
\begin{equation}
\mathcal{X}_i=\mathrm{RoIAlign}(S_i,\{r_{i,n}\})\in\mathbb{R}^{N_i\times h_r\times w_r\times C_s}.
\end{equation}
These region features undergo adaptive average pooling to a fixed spatial size ($7{\times}7$) followed by a lightweight convolutional head $g_\theta$, yielding compact descriptors:
\begin{equation}
\mathcal{Z}_i=g_\theta(\mathrm{AAPool}_{7\times7}(\mathcal{X}_i))\in\mathbb{R}^{N_i\times d},
\end{equation}
with centroids denoted as $K_i=\{k_{i,n}\in\mathbb{R}^2\}$.

We employ SuperGlue~\cite{sarlin2020superglue} to compute cross-epoch correspondence scores between region descriptors:
\begin{equation}
[Z_{pq}]_{N_1\times N_2} = \mathrm{SuperGlue}(K_1, \mathcal{Z}_1;\, K_2, \mathcal{Z}_2),
\end{equation}
where each entry $Z_{pq}$ represents the matching confidence between region $p$ from epoch~1 and region $q$ from epoch~2. Higher values indicate stronger geometric and semantic consistency. Reliable matches are determined by applying a threshold $\tau$:
\begin{equation}
Z_{pq} =
\begin{cases}
1, & \text{if } Z_{pq} > \tau,\\[4pt]
0, & \text{otherwise.}
\end{cases}
\end{equation}

Unmatched instances, potentially corresponding to changed or newly appeared/disappeared regions, are identified by:
\[
U_1=\{p\mid \sum_q Z_{pq}=0\},\quad U_2=\{q\mid \sum_p Z_{pq}=0\}.
\]
To incorporate spatial information, centroids of unmatched regions are embedded via positional encoding and concatenated with their respective region descriptors. 
The final motion-level change representation is obtained by concatenating these enriched features within each image, capturing structural changes at pixel-level precision without assuming object motion.



\subsubsection{Semantic-Level Change Aggregation}


To overcome the limitations of purely visual or manually defined prompts, we introduce Grounding DINO~\cite{liu2024grounding} as a semantic prior to automatically generate prompt-conditioned bounding boxes for both images $X_1$ and $X_2$. 
These boxes correspond to high-level semantic categories such as \textit{building}, \textit{road}, and \textit{vegetation}. 
Subsequently, SAM refines these bounding boxes into detailed segmentation masks, resulting in region-specific dense feature maps $S_i\in\mathbb{R}^{h\times w\times C_s}$.

Given textual prompts $\mathcal{T}=\{t_k\}_{k=1}^{K}$, we compute their text embeddings $E_T=\mathrm{enc}_T(\mathcal{T})\in\mathbb{R}^{K\times d}$. For each segmented region, we extract region embeddings $\mathcal{E}\in\mathbb{R}^{J\times d}$ using the same RoI pipeline described above. Image–text similarity is computed via normalized dot products:
\begin{equation}
S = \mathrm{Norm}(\mathcal{E})\,\mathrm{Norm}(E_T)^{\top} \in \mathbb{R}^{J\times K}.
\end{equation}
Regions are ranked based on their highest similarity scores across all prompts, and the top-$q$ regions are selected:
\begin{equation}
\mathcal{I}_{\mathrm{sel}} = \mathrm{Top}\text{-}q \big(\max_{k} S_{:,k}\big).
\end{equation}

The final semantic-level change representation is constructed by concatenating the embeddings of selected regions from each prompt category.
This semantic-aware representation effectively complements the motion-level change features, yielding a more coherent and interpretable representation of change across epochs.


\subsection{Remote-Sensing Change Graph Reasoner}
\label{sec:rscgr}

This module constructs a compact remote-sensing change knowledge graph from caption corpora and employs relation-aware message passing to derive semantic priors for subsequent reasoning and caption generation tasks. 

\begin{figure}
\centering
\includegraphics[width=0.45\textwidth]{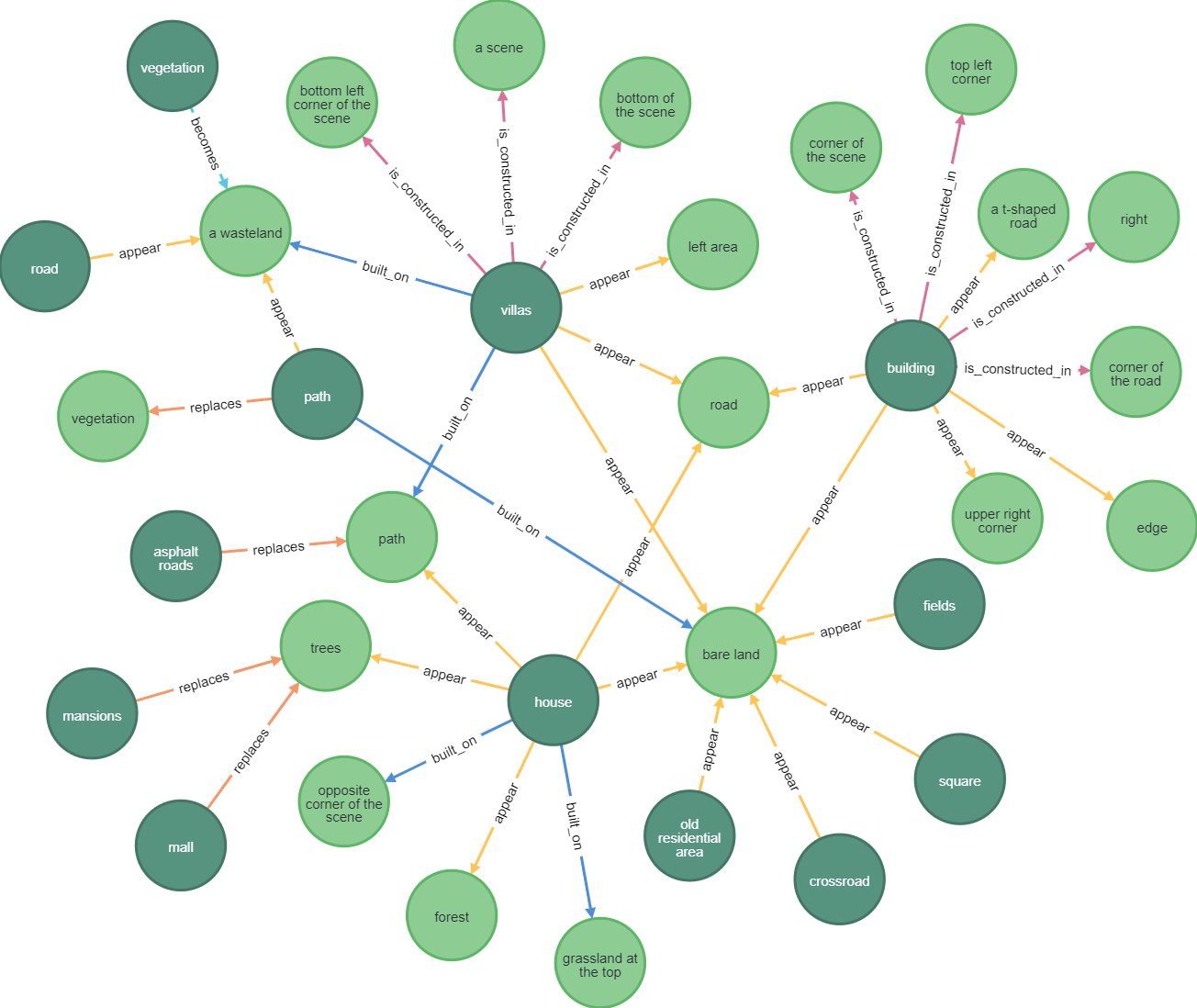}
\caption{Illustration of the built knowledge graph for the remote sensing change captioning task. Zoom in for better visualization.}   
\label{fig:graph} 
\end{figure}

\subsubsection{Graph construction from captions}
\label{sec:rscgr-build}


%
Given a set of descriptive captions $\mathcal{C}$ from multi-temporal remote-sensing datasets, each caption describes objects, land-cover types, and their transformations over time.
We apply a relation-extraction approach specifically adapted to remote-sensing change descriptions to extract structured semantic triples $\mathcal{T}\subseteq\mathcal{E}\times\mathcal{R}\times\mathcal{E}$, where $\mathcal{E}$ and $\mathcal{R}$ denote the raw entity and relation vocabularies. 
For instance, the caption ``a crossroad and several buildings appear on the bareland'' yields triples such as 
$\langle\,\text{\texttt{crossroad}},\,\text{\texttt{appear-on}},\,\text{\texttt{bareland}}\,\rangle$ and
$\langle\,\text{\texttt{building}},\,\text{\texttt{appear-on}},\,\text{\texttt{bareland}}\,\rangle$.


To mitigate redundancy and synonym overlaps common in natural-language annotations, we perform iterative clustering guided by a LLM~\cite{bi2024deepseek} with instruction set $\mathbf{I}_E$ to merge semantically similar entities:
\[
\mathcal{E}' = \mathrm{Merge}_{\mathbf{I}_E}(\mathcal{E}), \quad \pi:\mathcal{E}\rightarrow\mathcal{E}'.
\]
All triples are subsequently remapped via $\pi$ and deduplicated to produce the normalized knowledge graph:
\[
\mathcal{G}=(\mathcal{E}',\mathcal{R},\mathcal{T}').
\]

For computational efficiency, we encode the graph $\mathcal{G}$ using two matrices: 
the connectivity matrix
\begin{equation}
\label{eq:conn-type}
\begin{aligned}
A_{\mathrm{conn}} &=
\begin{bmatrix}
\mathrm{src}_1 & \mathrm{src}_2 & \cdots & \mathrm{src}_m\\
\mathrm{tgt}_1 & \mathrm{tgt}_2 & \cdots & \mathrm{tgt}_m
\end{bmatrix},\\[4pt]
A_{\mathrm{type}} &=
\begin{bmatrix}
r_1 & r_2 & \cdots & r_m
\end{bmatrix},
\end{aligned}
\end{equation}
where $m=|\mathcal{T}'|$, $\mathrm{src}_j,\mathrm{tgt}_j\in\{1,\dots,|\mathcal{E}'|\}$ denote indices of head and tail entities, and $r_j\in\{1,\dots,|\mathcal{R}|\}$ encodes relation types corresponding to land-use transitions and spatial interactions.

\begin{table*}
\centering
\caption{Comparisons with state-of-the-art RSICC methods on the LEVIR-CC dataset}
\label{tab:comparison LEVIR-CC}
\begin{tabular}{l|cccccccc}
\hline
Method & BLEU-1 & BLEU-2 & BLEU-3 & BLEU-4 & METEOR & ROUGE-L & CIDEr \\
\hline
Capt-Rep-Diff~\cite{park2019robust} & 72.90 & 61.98 & 53.62 & 47.41 & 34.47 & 65.64 & 110.57 \\
Capt-Att~\cite{park2019robust} & 77.64 & 67.40 & 59.24 & 53.15 & 36.58 & 69.73 & 121.22 \\
Capt-Dual-Att~\cite{park2019robust} & 79.51 & 67.23 & 57.46 & 36.56 & 37.16 & 69.19 & 124.42 \\
DUDA~\cite{park2019robust} & 81.44 & 72.22 & 64.67 & 57.79 & 37.15 & 71.04 & 124.32 \\
MCCFormer-S~\cite{qiu2021describing} & 79.90 & 70.26 & 62.68 & 56.36 & 39.60 & 69.46 & 120.39  \\
MCCFormer-D~\cite{qiu2021describing} & 80.42 & 70.87 & 62.86 & 56.38 & 39.91 & 70.44 & 124.44  \\
RSICCformer-C~\cite{liu2022remote} & 83.09 & 74.32 & 66.66 & 62.41 & 38.70 & 73.60 & 132.62 \\
PSNet~\cite{liu2023progressive} & 83.86 & 75.13 & 67.89 & 62.11 & 38.80 & 73.60 & 132.62 \\
Chg2Cap~\cite{chang2023changes} & 84.43 & 76.35 & 69.12 & 62.98 & 39.42 & 74.34 & 136.25 \\
SEN~\cite{zhou2024single}& 85.10 & 77.05 & 70.01 & 64.09 & 39.59 & 74.57 & 136.02 \\
Diffusion-RSCC~\cite{yu2025diffusion}& - & - & - & 60.90 & 37.80 & 71.50 & 125.60 \\
CTMTNet~\cite{shi2024multi}& \textbf{85.95} & \textbf{77.99} & 70.74 & 64.69 & 39.49 & 74.54 & 134.94\\
\hline
\textbf{Ours} & 85.69 & 77.85 & \textbf{71.03} & \textbf{65.50} & \textbf{39.92} & \textbf{74.77} & \textbf{137.50} \\
\hline
\end{tabular}
\end{table*}

\begin{table*}
\centering
\caption{Comparisons with state-of-the-art RSICC methods on the Dubai-CC dataset}
\label{tab:comparison DUBAI-CC}
\begin{tabular}{l|cccccccc}
\hline
Method & BLEU-1 & BLEU-2 & BLEU-3 & BLEU-4 & METEOR & ROUGE-L & CIDEr-D  \\
\hline
DUDA~\cite{park2019robust} & 58.82 & 43.59 & 33.63 & 25.39 & 22.05 & 48.34 & 62.78  \\
MCCFormer-S~\cite{qiu2021describing} & 52.97 & 37.02 & 27.62 & 22.57 & 18.64 & 43.29 & 53.81  \\
MCCFormer-D~\cite{qiu2021describing} & 64.65 & 50.45 & 39.36 & 29.48 & 25.09 & 51.27 & 63.09  \\
RSICCFormer-C~\cite{liu2022remote} & 67.92 & 53.61 & 41.37 & 31.28 & 25.41 & 51.96 & 66.54  \\
Diffusion-RSCC~\cite{yu2025diffusion}& - & - & - & 33.30 & 27.40 & 56.50 & 88.70 \\
Chg2Cap~\cite{chang2023changes} & 72.04 & 60.18 & 50.84 & 41.70 & 28.92 & 58.66 & 92.49  \\
\hline
\textbf{Ours} & \textbf{74.25} & \textbf{62.12} & \textbf{51.77} & \textbf{42.21} & \textbf{29.05} & \textbf{59.58} & \textbf{93.26}  \\
\hline
\end{tabular}
\end{table*}

\subsubsection{Relation-aware graph encoding}
\label{sec:rscgr-encode}

Entities are initialized using contextualized text embeddings. Let $e'_i\in\mathcal{E}'$ denote the $i$-th entity; a pretrained text encoder (e.g., BERT) generates initial node embeddings:
\begin{equation}
V=\mathrm{BERT}([e'_1,\dots,e'_{|\mathcal{E}'|}])\in\mathbb{R}^{|\mathcal{E}'|\times d_b},
\end{equation}
with $h_i^{(0)}=V_i$. 

We employ a multi-layer Relational Graph Convolutional Network (R-GCN)~\cite{schlichtkrull2018modeling} to propagate and aggregate relation-specific features across graph nodes:
\begin{equation}
\label{eq:rgcn}
h_i^{(l+1)}=\sigma\left(
W_0^{(l)}h_i^{(l)} +
\sum_{r\in\mathcal{R}}\sum_{j\in\mathcal{N}_i^{r}}
\frac{1}{c_{i,r}}W_r^{(l)}h_j^{(l)}
\right),
\end{equation}
where $\mathcal{N}_i^{r}$ denotes neighboring nodes of node $i$ linked by relation type $r$ (retrievable from $A_{\mathrm{conn}}$, $A_{\mathrm{type}}$), $c_{i,r}$ is a normalization constant, $W_r^{(l)}$ and $W_0^{(l)}$ are relation-specific and self-loop weights, and $\sigma$ is an activation function.

Unlike standard GCN architectures, the R-GCN explicitly conditions message-passing transformations on relation type and edge directionality, enabling it to encode asymmetric dependencies. 
This is particularly crucial for remote sensing tasks where relationships like ``\texttt{bareland→building}'' differ fundamentally from ``\texttt{building→bareland}'' in semantic meaning and temporal direction.

Final node embeddings are aggregated into a graph-level representation using a permutation-invariant readout operation:
\begin{equation}
F_{\mathrm{kg}}=\mathrm{READOUT}\!\big(H^{(L)}\big)\in\mathbb{R}^{d_{\mathrm{kg}}},
\end{equation}
where $H^{(L)}=[h_1^{(L)};\dots;h_{|\mathcal{E}'|}^{(L)}]$, and the READOUT function denotes mean or attention-based pooling, optionally conditioned on task-specific prompts.

The resultant embedding $F_{\mathrm{kg}}$ encapsulates critical entity dependencies and prevalent change patterns, 
serving as a semantic prior to guide downstream caption generation and reasoning modules.

\begin{table*}
\centering
\caption{Comparisons with state-of-the-art RSICC methods on the WHU-CDC dataset}
\label{tab:comparison WHU-CDC}
\begin{tabular}{l|cccccccc}
\hline
Method & BLEU-1 & BLEU-2 & BLEU-3 & BLEU-4 & METEOR & ROUGE-L & CIDEr-D  \\
\hline
DUDA~\cite{park2019robust} & 79.04 & 69.53 & 61.57 & 55.64 & 34.29 & 68.98 & 121.85  \\
MCCFormer-S~\cite{qiu2021describing} & 82.14 & 76.29 & 71.08 & 66.51 & 43.50 & 79.76 & 148.88  \\
MCCFormer-D~\cite{qiu2021describing} & 73.29 & 67.88 & 64.03 & 60.96 & 39.69 & 73.67 & 134.92  \\
RSICCformer-C~\cite{liu2022remote} & 78.25 & 72.82 & 68.57 & 65.14 & 44.35 & 76.50 & 143.44 \\
MaskApproxNet~\cite{sun2025mask} & 81.34 & 75.68 & 71.16 &  67.73 & 43.89 & 75.41 & 135.31  \\
CTMTNet~\cite{shi2024multi} & 83.56 & 77.66 & 72.76 & 69.00 & 45.39 & 79.23 & 149.40 \\
CTM~\cite{bai2025cross} & 85.36 & 79.49 & 75.36 & 72.36 & 46.98 & \textbf{80.97} & 153.29 \\
\hline
\textbf{Ours} & \textbf{86.04} & \textbf{81.16} & \textbf{77.28} & \textbf{74.42} & \textbf{47.89} & 80.90 & \textbf{156.21}  \\
\hline
\end{tabular}
\end{table*}

\begin{table*}
\centering
\caption{Ablation study on the LEVIR-CC dataset, assessing the impact of key components:Motion-Level Change Localization (MCL), SuperGlue (SG), Semantic-Level Change Aggregation (SCA), and Remote-Sensing Change Graph Reasoner (CGR). Symbols ``$\times$'' and ``$\checkmark$'' indicate the exclusion and inclusion of specific modules, respectively. Higher metric values denote better performance, with best results highlighted in bold.}
\label{tab:ablation_all}
\begin{tabular}{l|cccc|ccccccc}
\hline
Method & MCL & SG & SCA & CGR & BLEU-1 & BLEU-2 & BLEU-3 & BLEU-4 & METEOR & ROUGE-L & CIDEr-D \\
\hline
Baseline & $\times$ & $\times$ & $\times$ & $\times$ & 84.43 & 76.35 & 69.12 & 62.98 & 39.42 & 74.34 & 136.25 \\
(a) & $\checkmark$ & $\times$ & $\times$ & $\times$ & 85.05 & 76.59 & 69.50 & 63.66 & 39.71 & {74.63} & 137.01 \\
(b) & $\checkmark$ & $\checkmark$ & $\times$ & $\times$ & 84.99 & 76.72 & 69.81 & 64.08 & 39.60 & 74.09 & 136.12 \\
(c) & $\checkmark$ & $\checkmark$ & $\checkmark$ & $\times$ & \textbf{85.82} & {77.66} & {70.46} & {64.47} & {39.83} & 74.66 & \textbf{138.01} \\
\hline
(d) & $\checkmark$ & $\checkmark$ & $\checkmark$ & $\checkmark$ &  {85.69} & \textbf{77.85} & \textbf{71.03} & \textbf{65.50} & \textbf{39.92} & \textbf{74.77} & {137.50} \\
\hline
\end{tabular}
\end{table*}

\subsection{Change-Aware Language Generator}
\label{sec:calg}

\noindent\textbf{Change-Conditioned Fusion and Generation.}
We first integrate the motion-level change representation $F_{\mathrm{disc}}\in\mathbb{R}^{d'}$, the semantic-level change representation $F_{\mathrm{sem}}\in\mathbb{R}^{d}$, and the graph-derived semantic prior $F_{\mathrm{kg}}\in\mathbb{R}^{d_{\mathrm{kg}}}$ through linear projections and concatenation:
\begin{equation}
r = \mathrm{LN}\bigl([W_d F_{\mathrm{disc}};\, W_s F_{\mathrm{sem}};\, W_g F_{\mathrm{kg}}]\bigr)\in\mathbb{R}^{d_f}.
\end{equation}

Given the global embedding $E_{\mathrm{img}}\in\mathbb{R}^{h\times w\times C_e}$, we flatten it into token embeddings $X\in\mathbb{R}^{(hw)\times d_e}$ to serve as cross-attention inputs for the decoder. 
The decoder consists of $L$ Transformer layers, each incorporating masked self-attention over caption tokens and cross-attention to visual tokens $X$. 
To explicitly guide cross-attention towards changed regions, we inject a spatial attention bias using consistency priors from Sec.~\ref{sec:btsce}:
\begin{equation}
\begin{aligned}
\pi &= \alpha\,\mathrm{vec}(M_{\mathrm{map}}) + \beta\,\mathrm{vec}(C_{\mathrm{map}})\in \mathbb{R}^{hw},\\[4pt]
B &= \mathbf{1}_{T}\,\pi^{\top}\in \mathbb{R}^{T\times(hw)},
\end{aligned}
\end{equation}
where $\alpha,\beta$ are learnable scaling parameters.

Let $Y^{(0)}\in\mathbb{R}^{T\times d_f}$ represent the initial caption token embeddings. 
Each Transformer decoder layer $\ell$ computes:
\begin{align}
\widetilde Y^{(\ell)} &= \mathrm{SelfAttn}\big(Y^{(\ell-1)}\big),\\[4pt]
Z^{(\ell)} &= \mathrm{CrossAttn}\big(\widetilde Y^{(\ell)},\,K{=}X,\,V{=}X;\,B\big),\\[4pt]
Y^{(\ell)} &= \mathrm{FFN}\bigl(\mathrm{LN}(\widetilde Y^{(\ell)} + Z^{(\ell)} + \mathbf{1}_T r^{\top})\bigr),
\end{align}
where $\mathbf{1}_T r^{\top}$ explicitly integrates the fused change-aware representation $r$ into each decoding step. 
Finally, the hidden states $H=Y^{(L)}$ of the last decoder layer are projected onto vocabulary logits and probabilities:
\begin{equation}
\mathrm{logits} = W_o H + b_o,\quad P=\mathrm{Softmax}(\mathrm{logits}).
\end{equation}

\noindent\textbf{Loss Function.}
We train the decoder via teacher forcing with the standard cross-entropy (CE) objective against target caption tokens $\{y_t^\ast\}_{t=1}^{T}$:
\begin{equation}
\mathcal{L}_{\mathrm{CE}}
= -\sum_{t=1}^{T} \log P\big(y_t^\ast \mid y_{<t}, X, r, B\big).
\end{equation}
To improve model generalization, we optionally employ label smoothing with factor $\varepsilon$: each ground-truth token is represented as $(1-\varepsilon)$ for the correct word and $\varepsilon/(V-1)$ evenly distributed across remaining vocabulary tokens, where $V$ denotes the vocabulary size.

\section{Experiments} 
\begin{table*}
\centering
\caption{Ablation study on the number of masks in the MCL module on the LEVIR-CC dataset.}
\begin{tabular}{l|cccccccc}
\hline
Number & BLEU-1 & BLEU-2 & BLEU-3 & BLEU-4 & METEOR & ROUGE-L & CIDEr \\
\hline
30 & 85.13 & 76.98 & 69.83 & 63.74 & 39.81 & 74.60 & 137.09 \\
40 & 85.31 & 77.07 & 70.15 & \textbf{64.48}& 39.52 & 74.32 & 136.63 \\
50 & \textbf{85.82} & \textbf{77.66} & \textbf{70.46} & 64.47 & \textbf{39.83} & \textbf{74.66} & \textbf{138.01} \\
\hline
\end{tabular}
\label{tab:mask_comparison}
\end{table*}

\begin{table*}
\centering
\caption{Ablation study on the threshold of SG on the LEVIR-CC dataset.}
\begin{tabular}{l|cccccccc}
\hline
Threshold & BLEU-1 & BLEU-2 & BLEU-3 & BLEU-4 & METEOR & ROUGE-L & CIDEr \\
\hline
0.1 & 84.82 & 76.70 & 69.57 & 63.57 & 39.71 & 74.35 & 137.13 \\
0.2 & \textbf{85.82} & \textbf{77.66} & \textbf{70.46} & \textbf{64.47} & \textbf{39.83} & \textbf{74.66} & \textbf{138.01} \\
0.3 & 85.03 & 76.94 & 69.73 & 63.87 & 39.48 & 73.88 & 134.39 \\
\hline
\end{tabular}
\label{tab:threshold_comparison}
\end{table*}

\begin{table*}
\centering
\caption{Ablation study on different graph encoders for model performance on the LEVIR-CC dataset.}
\begin{tabular}{l|cccccccc}
\hline
Encoder & BLEU-1 & BLEU-2 & BLEU-3 & BLEU-4 & METEOR & ROUGE-L & CIDEr \\
\hline
GCN & 85.23 & 77.01 & 69.73 & 63.80 & 39.56 & 74.17 & 135.62 \\
GAT & 85.42 & 77.20 & 70.21 & 64.51 & 39.59 & 74.37 & 136.20 \\
SAGE & 84.06 & 75.91 & 68.87 & 63.10 & 39.26 & 73.79 & 134.80 \\
MONET & 82.71 & 74.15 & 67.16 & 61.64 & 38.22 & 72.58 & 130.20 \\
RGCN & \textbf{85.69} & \textbf{77.85} & \textbf{71.03} & \textbf{65.50} & \textbf{39.92} & \textbf{74.77} & \textbf{137.50} \\

\hline
\end{tabular}
\label{tab:encoder_comparison}
\end{table*}

\begin{table*}
\centering
\caption{Ablation study on the frequency threshold $k$ for filtering low-frequency relationships and entities on the LEVIR-CC dataset.}
\begin{tabular}{l|cccccccc}
\hline
$k$ & BLEU-1 & BLEU-2 & BLEU-3 & BLEU-4 & METEOR & ROUGE-L & CIDEr \\
\hline
30 & 84.21 & 76.15 & 69.26 & 63.50 & 39.23 & 73.83 & 134.90 \\
40 & 85.10 & 77.25 & 70.29 & 64.58 & 39.82 & 74.55 & 135.49 \\
50 & \textbf{85.69} & \textbf{77.85} & \textbf{71.03} & \textbf{65.50} & 39.92 & 74.77 & 137.50 \\
60 & 85.22 & 77.41 & 70.45 & 64.63 & \textbf{40.28} & \textbf{75.23} & \textbf{138.21} \\
70 & 85.18 & 77.06 & 69.92 & 64.14 & 39.78 & 74.39 & 135.69 \\

\hline
\end{tabular}
\label{tab:k_comparison}
\end{table*}

\subsection{Datasets and Evaluation Metric}  

\noindent $\bullet$ \textbf{LEVIR-CC} 
The LEVIR-CC dataset~\cite{liu2022remote} comprises 10,077 pairs of bitemporal remote sensing images (5,038 with changes and 5,039 without changes) derived from the LEVIR-CD dataset~\cite{chen2020spatial}. Each image has a 256 × 256 pixel size with 0.5 m/pixel resolution, acquired from 20 regions across Texas via Google Earth API with a 5-14 year time span between acquisitions. Each image pair is annotated with five descriptive sentences (50,385 total captions), with fixed sentences for non-change pairs and varied descriptions for change pairs. Following the default experimental settings~\cite{liu2022remote}, we split the dataset into 6,815 pairs for training, 1,333 for validation, and 1,929 for testing.

\noindent $\bullet$ \textbf{Dubai-CC} 
The Dubai-CC dataset~\cite{hoxha2022change} consists of 500 pairs of bitemporal remote sensing images capturing urbanization changes in the Dubai area. The images were acquired by the Enhanced Thematic Mapper Plus (ETM+) sensor onboard Landsat 7 on May 19, 2000, and June 16, 2010. The original images were cropped into 50 × 50 pixel tiles, with each pair annotated with five different change descriptions referencing Google Maps and publicly available documents, resulting in 2,500 independent descriptions. Following the default experimental settings~\cite{hoxha2022change}, we split the dataset into 300 pairs for training, 50 for validation, and 150 for testing.

\noindent $\bullet$ \textbf{WHU-CDC}
The WHU-CDC dataset~\cite{shi2024multi} contains 7,434 high-resolution bi-temporal image pairs spanning from 2011 to 2016, which describe changes in buildings, parking lots, roads, and other categories. In total, the dataset provides 37,170 descriptive sentences. Following the default experimental settings~\cite{shi2024multi}, we split the dataset into 5947 pairs for training, 743 for validation, and 744 for testing.

\noindent $\bullet$ \textbf{Evaluation Metric}
BLEU-N measures the n-gram precision between the generated sentences and the reference sentences, emphasizing local lexical overlap. ROUGE-L computes recall based on the longest common subsequence, reflecting the extent to which the generated text covers the content of the references. METEOR takes both precision and recall into account, incorporating stemming, synonym matching, and a fluency penalty, thereby better capturing semantic accuracy. CIDEr-D, specifically designed for image captioning tasks, employs TF-IDF weighting for n-grams to highlight rare yet informative phrases.

\subsection{Implementation Details} 
The deep learning methods proposed in this study are implemented using the PyTorch~\cite{paszke2019pytorch} framework, with all model training and evaluation conducted on an NVIDIA RTX 4090 GPU equipped with 24 GB of memory. During training, the Adam optimizer is employed with an initial learning rate of 0.0001, which is decayed by a factor of 0.5 after 5 epochs. The maximum number of training epochs is set to 50. After each epoch, the model is validated on the development set, and the checkpoint achieving the highest BLEU-4 score is retained as the final model for evaluation on the test set. More details can be found in our \textit{source code}.

\subsection{Comparison on Public Benchmarks} 
As shown in Table~\ref{tab:comparison LEVIR-CC}, Table~\ref{tab:comparison DUBAI-CC} and Table~\ref{tab:comparison WHU-CDC}, we conducted comparisons with several existing state-of-the-art methods on three key remote sensing change captioning datasets (LEVIR-CC, Dubai-CC, and WHU-CDC). The performance of our method, SAGE-CC, was evaluated using multiple evaluation metrics, including BLEU-1, BLEU-2, BLEU-3, BLEU-4, METEOR, ROUGE-L, and CIDEr.

On the LEVIR-CC dataset, SAGE-CC ranks among the top performers across multiple metrics, with BLEU-3 at 71.03, BLEU-4 at 65.50, ROUGE-L at 39.92, and METEOR at 74.77. Additionally, the CIDEr score reaches 137.50, demonstrating the effectiveness of our method in generating descriptions highly consistent with human annotations. On the Dubai-CC dataset, SAGE-CC also exhibits strong performance, achieving excellent results across all evaluation metrics. The BLEU-1 score is 74.25, and the BLEU-4 score is 42.21. Furthermore, the CIDEr score of 93.26 further highlights the advantage of our model in generating high-quality descriptions. On the WHU-CDC dataset, SAGE-CC again leads with a BLEU-1 score of 86.04 and a BLEU-4 score of 74.42, with a CIDEr score as high as 156.21. These results further validate the superiority of our method in generating precise and detailed remote sensing image descriptions.

\subsection{Ablation Study} 
As shown in Table~\ref{tab:ablation_all}, we systematically evaluated the contributions of four key components, Motion-Level Change Localization (MCL, without SG), SuperGlue (SG), Semantic-Level Change Aggregation (SCA), and Remote-Sensing Change Graph Reasoner (CGR), to the overall performance of the model on the LEVIR-CC dataset. The results in the table indicate that, compared to the baseline model (Baseline), the performance on natural language generation metrics (BLEU-1 to BLEU-4, METEOR, ROUGE-L, CIDEr-D) improves to varying degrees as each module is progressively added. This demonstrates that each component effectively enhances the model's ability to capture and describe changed regions. Specifically, Method (a), after incorporating MCL, shows an improvement in BLEU-4 from 62.98 to 63.66 and CIDEr-D from 136.25 to 137.01, indicating that motion-level changed regions help generate more precise change information. Method (b), after integrating SG, shows improvements across all metrics, proving the effectiveness of the feature matching module. Method (c), with the further addition of SCA, increases BLEU-4 from 64.08 to 64.47, suggesting that the semantic-level assistance plays a positive role in the accuracy of change detection. Finally, Method (d), which introduces the CGR module, increases BLEU-4 to 65.50, maintaining high performance while balancing BLEU and CIDEr-D, demonstrating the robustness of the overall model.

In summary, the experimental results fully validate the effectiveness of each proposed module in improving model performance. The combination of MCL and SG plays a key role in fine-grained change modeling, SCA provides valuable support in the semantic aspect, and the CGR module, through the extraction of structured information, enhances the accuracy and consistency of remote sensing change descriptions.

\begin{figure*}[!htp]
\centering
\includegraphics[width=1\textwidth]{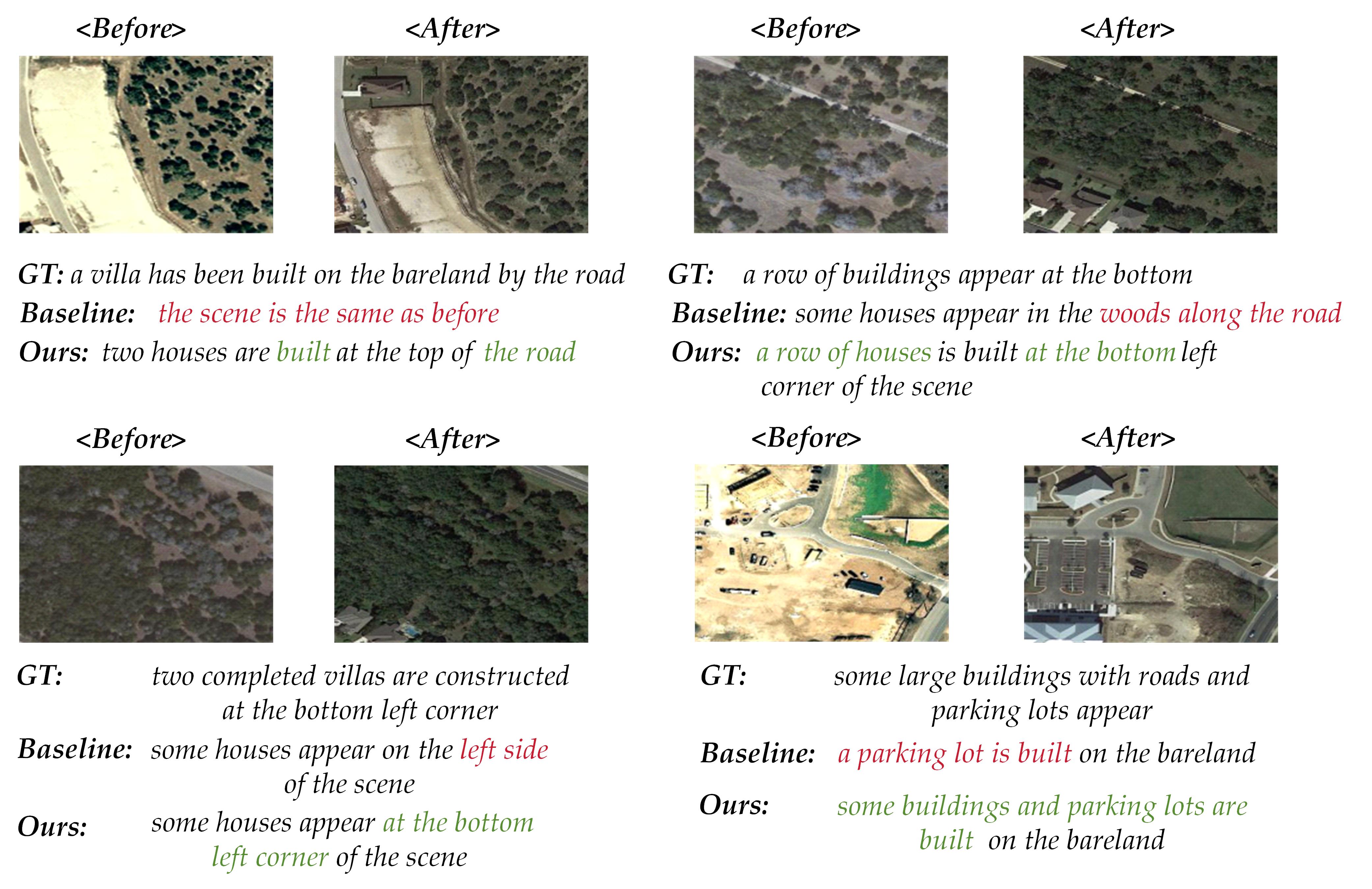}
\caption{Remote sensing change image pairs and their corresponding ground-truth annotations are provided in the LEVIR-CC dataset. The matching parts are shown in green, and the mismatched parts are shown in red.} 
\label{fig:captioning results} 
\end{figure*}

\subsection{Parameter Analysis} 

\noindent $\bullet$ \textbf{Analysis of Numbers of Masks in MCL Module.}
As shown in Table~\ref{tab:mask_comparison}, we investigated the impact of different numbers of masks in the MCL module on the model's performance. The results indicate that as the number of masks increases from 30 to 50, the overall performance of the model gradually improves. Specifically, when the number of masks is 50, the model achieves optimal performance in metrics such as BLEU-1, BLEU-2, BLEU-3, and CIDEr, with BLEU-4 and CIDEr reaching 64.47 and 138.01, respectively. This suggests that increasing the number of masks appropriately allows for a more comprehensive coverage of motion-level changed regions, thereby providing richer fine-grained features for change detection. When the number of masks is too small, the model struggles to capture sufficient change information, leading to a decrease in generation performance.

\noindent $\bullet$ \textbf{Analysis of Matching Threshold in SG Module.}
As shown in Table~\ref{tab:threshold_comparison}, we further explored the impact of different matching threshold settings in the SG module on the model's performance. The experimental results demonstrate that when the threshold is set to 0.2, the model reaches a peak CIDEr score of 138.01, while BLEU-4 and ROUGE-L also maintain their highest levels. A threshold that is too low (0.1) or too high (0.3) leads to a performance decline. The former may introduce excessive irrelevant information, while the latter may overlook valid correspondences. Therefore, setting an appropriate threshold is crucial for balancing robustness and precision.

\noindent $\bullet$ \textbf{Comparison of Graph Encoding Methods.}
As shown in Table~\ref{tab:encoder_comparison}, we compared the performance of different graph encoding methods (GCN, GAT, SAGE, MONET, and RGCN) on BLEU, ROUGE-L, METEOR, and CIDEr metrics. The results show that RGCN achieves the best performance across all metrics, with BLEU-4, ROUGE-L, and CIDEr reaching 65.50, 74.77, and 137.50, respectively, which are significantly superior to other methods. In contrast, methods such as GCN exhibit relatively weaker performance, indicating their limitations in handling complex relational dependencies. We speculate that RGCN can explicitly model multiple relation types, thereby better capturing the dependency structures between different change regions and semantic units. This plays a crucial role in generating more coherent and accurate natural language descriptions.

\noindent $\bullet$ \textbf{Analysis of Threshold k for Filtering Low-Frequency Relations and Entities.}
As shown in Table~\ref{tab:k_comparison}, we further investigated the impact of the threshold \( k \) for filtering low-frequency relations and entities on the model's performance. The experimental results demonstrate that as \( k \) increases from 30 to 50, the model's performance generally improves. When \( k=50 \), the model achieves the best performance in BLEU, with BLEU-1, BLEU-2, BLEU-3, and BLEU-4 reaching 85.69, 77.85, 71.03, and 65.50, respectively, significantly outperforming other settings. When \( k \) is further increased to 60, BLEU performance shows a slight decline, but ROUGE-L, METEOR, and CIDEr improve to 75.23, 40.28, and 138.21, respectively. As \( k \) continues to increase, all performance metrics decline, suggesting that an excessively high threshold may overly filter out some valid relations and entities, thus weakening the model's expressive ability. In summary, \( k=50 \) is the optimal setting in this experiment, effectively balancing the retention of semantic richness while filtering out low-frequency irrelevant information.

\begin{figure*}
\centering
\includegraphics[width=\textwidth]{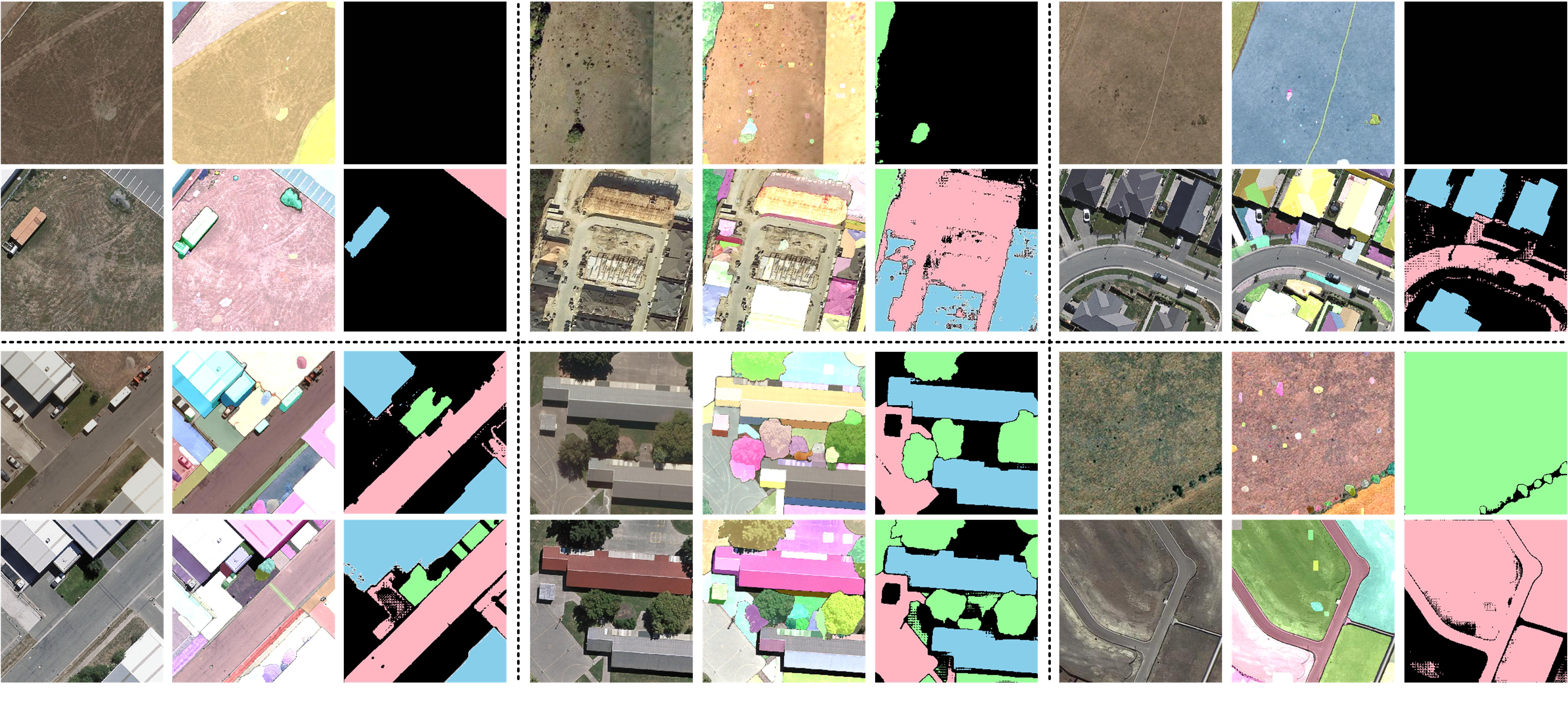}
\caption{Semantic segmentation results obtained using the SAM model. Blue, green, and pink regions denote building, vegetation, and road, respectively.} 
\label{fig:samvis} 
\end{figure*}

\begin{figure}
\centering
\includegraphics[width=\linewidth]{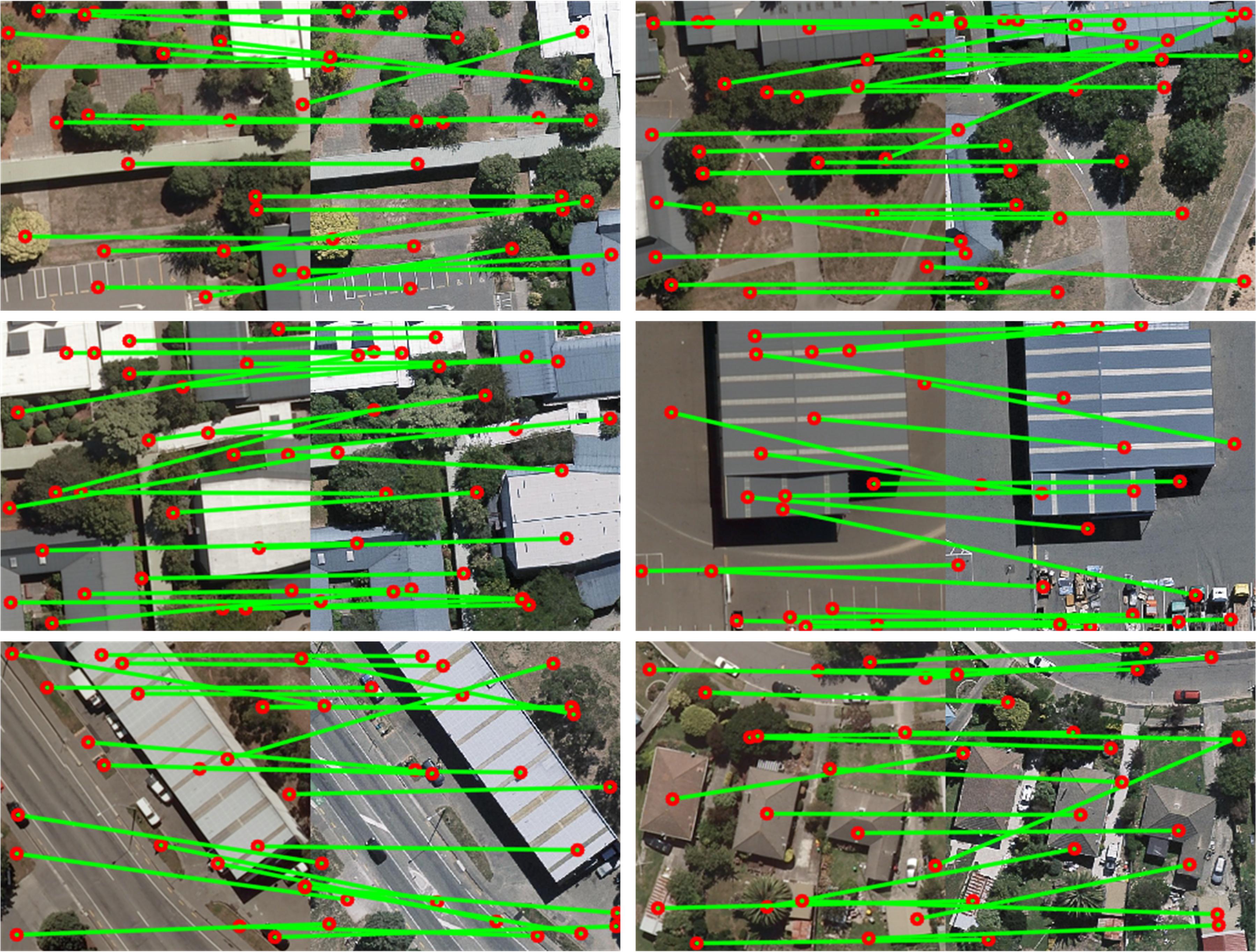}
\caption{Visualization of the matching results by SuperGlue.}  
\label{fig:superglue_vis} 
\end{figure}

\subsection{Visualization} 

\noindent $\bullet$ \textbf{Captioning Results.~} As shown in Figure~\ref{fig:captioning results}, we present some examples to demonstrate the effectiveness of our proposed SAGE-CC model in remote sensing change detection captioning. For specific dual-temporal remote sensing images, we compare the captions generated by the SAGE-CC model with those generated by the baseline model and the ground truth labels. To provide a more intuitive visualization, we use red font to indicate the inaccurate descriptions generated by the baseline model, while green font represents the portions of the captions generated by the SAGE-CC model that are consistent with the ground truth. From the visualization results, it is evident that the captions generated by our model are of higher quality and more consistent with the ground truth compared to those generated by the baseline model.

\noindent $\bullet$ \textbf{Visualization of Semantic Segmentation Masks.~} As shown in Figure~\ref{fig:samvis}, we present the semantic segmentation results obtained using the SAM on dual-temporal remote sensing images. Since there are many semantic categories but limited display space, we select the three most frequently appearing classes for visualization. In the segmentation maps, blue regions represent buildings, green regions indicate vegetation, and pink regions correspond to roads. By applying SAM-based segmentation to both temporal images, fine-grained object boundaries and class-level changes can be effectively captured, providing rich spatial and structural information for subsequent change captioning tasks.

\noindent $\bullet$ \textbf{Knowledge Graph Constructed from Captions.~}
As shown in Figure~\ref{fig:graph}, due to the large number of nodes, we only display a portion of the relationships between the nodes to ensure the clarity and readability of the graph. Each node in the figure represents an entity, and each arrow indicates a relationship between entities, with different arrow colors representing different types of relationships. This approach allows for a more intuitive visualization of the key information and interrelationships in remote sensing captions, helping to understand the complex patterns and potential connections within the captions.

\noindent $\bullet$ \textbf{Visualization of SuperGlue.~} As shown in Figure~\ref{fig:superglue_vis}, we present some matching results generated by the SuperGlue algorithm. The red points represent the mask feature points extracted from the two images, while the green lines indicate the successfully matched feature point pairs, demonstrating the spatial correspondence between the two images. These matches provide crucial support for subsequent change detection tasks.

\subsection{Limitation Analysis}  
The overall framework of SAGE-CC consists of multiple modules, aiming to achieve accurate change description. However, there are significant differences in granularity and semantic space between remote sensing change region features and text-based graphs, which may lead to a lack of deep alignment mechanisms in terms of structural hierarchy and semantic representation. This can affect the model’s performance in cross-modal fusion. Additionally, while the model processes fine-grained details through multiple modules, its overall structure is complex and computationally expensive, which may limit its deployment and application in resource-constrained scenarios.

\section{Conclusion}  
In this paper, we propose a novel approach to remote sensing change captioning that leverages SAM to extract region-level representations and integrates them with auxiliary knowledge sources to enhance change description. By fusing visual and semantic information through a Transformer architecture, we generate natural language descriptions with high geometric accuracy and semantic richness. Extensive experimental results demonstrate that our method outperforms existing approaches across multiple public benchmark datasets.

{
    \small
    \bibliographystyle{ieeenat_fullname}
    \bibliography{main}
}


\end{document}